\documentclass[10pt,journal,compsoc]{IEEEtran}
\IEEEoverridecommandlockouts
\ifCLASSOPTIONcompsoc
  \usepackage[nocompress]{cite}
\else
  \usepackage{cite}
\fi
\usepackage{amsmath,amssymb,amsfonts}
\usepackage{algorithmic}
\usepackage{graphicx}
\usepackage{textcomp}
\usepackage{xcolor}
\def\BibTeX{{\rm B\kern-.05em{\sc i\kern-.025em b}\kern-.08em
    T\kern-.1667em\lower.7ex\hbox{E}\kern-.125emX}}
    
\usepackage{caption}
\usepackage{subcaption}
\usepackage{xspace}
\usepackage{algorithm}
\usepackage{algorithmic}
\usepackage{balance}
\usepackage{float}

\usepackage{titlesec}

\usepackage[hyphens,spaces,obeyspaces]{url}
\urldef{\codeurl}\url{https://drive.google.com/drive/folders/1vPqsLW59wJqhMacB6HgSmMdFda6S3DdO}
\usepackage{hyperref}

\usepackage[mathscr]{euscript}
\DeclareSymbolFont{rsfs}{U}{rsfs}{m}{n}
\DeclareSymbolFontAlphabet{\mathscrsfs}{rsfs}

\usepackage{soul}

\newcommand\methodname{HMGE\xspace}

\begin{document}

\title{Hierarchical Aggregations for \\ High-Dimensional Multiplex Graph Embedding
}

\author{Kamel~Abdous,~
        Nairouz~Mrabah,~
        Mohamed~Bouguessa
\IEEEcompsocitemizethanks{\IEEEcompsocthanksitem K. Abdous, N. Mrabah, M. Bouguessa are with the Department
of Computer Science, University of Quebec at Montreal, Montreal, QC, Canada.
E-mails: abdous.kamel@courrier.uqam.ca, ~mrabah.nairouz@courrier.uqam.ca, ~bouguessa.mohamed@uqam.ca}
}

\markboth{Journal of \LaTeX\ Class Files,~Vol.~x, No.~x, Month~Year}%
{Shell \MakeLowercase{\textit{et al.}}: Bare Demo of IEEEtran.cls for Computer Society Journals}


\IEEEtitleabstractindextext{
\begin{abstract}

We investigate the problem of multiplex graph embedding, that is, graphs in which nodes interact through multiple types of relations (dimensions). In recent years, several methods have been developed to address this problem. However, the need for more effective and specialized approaches grows with the production of graph data with diverse characteristics. In particular, real-world multiplex graphs may exhibit a high number of dimensions, making it difficult to construct a single consensus representation. Furthermore, important information can be hidden in complex latent structures scattered in multiple dimensions. To address these issues, we propose \methodname, a novel embedding method based on hierarchical aggregation for high-dimensional multiplex graphs. Hierarchical aggregation consists in learning a hierarchical combination of the graph dimensions and refining the embeddings at each hierarchy level. Non-linear combinations are computed from previous ones, thus uncovering complex information and latent structures hidden in the multiplex graph dimensions. Moreover, we leverage mutual information maximization between local patches and global summaries to train the model without supervision. This allows to captures globally relevant information present in diverse locations of the graph. Detailed experiments on synthetic and real-world data illustrate the suitability of our approach on downstream supervised tasks, including link prediction and node classification. 


\end{abstract}

}

\maketitle

\IEEEdisplaynontitleabstractindextext

\IEEEpeerreviewmaketitle


\IEEEraisesectionheading{\section{Introduction}}

\subsection{Context and Background Information}

\IEEEPARstart{T}{oday’s} real systems are mostly made of entities that interact with each other through multiple channels of connectivity. To better represent such complex systems, a new class of graphs, called \textit{multiplex graphs}, has emerged. In a multiplex graph, nodes (entities) are connected to each other through multiple types of relations (links) \cite{kivela2014multilayer}. 
We refer to these relations as dimensions. 
Such graphs arise in various areas, such as protein-protein interactions \cite{oughtred2021biogrid}, neuroimaging \cite{zhang2018multi}, social networks \cite{berlingerio2013multidimensional}, online recommendation \cite{sun2019multi}. In general, multiplexes provide a more general framework that allows rich and flexible modeling of several interconnected systems.





In recent years, many efforts have been made to design effective mining algorithms for multiplex graphs \cite{sanchez2014dimensionality, de2015structural, sole2016random, el2020orthonet, fan2020one2multi}. Most of these approaches are based on graph embedding techniques, which aim to project the graph dimensions 
to a compact and low-dimensional latent space representation. An effective embedding method should achieve exploitable representations for various prediction tasks, such as node classification \cite{park2020dmgi, jing2021hdmi} and link prediction \cite{chu2019cross, pio2021multiverse}. 
This is not straightforward because multiplex graphs may have a large number of dimensions, each containing various complementary and / or divergent information on node interactions \cite{de2015structural}. 
Therefore, an optimal embedding method should only encode the intrinsic informative structures hidden in the graph dimensions.


Over the past few years, a number of embedding methods for multiplex graphs have been proposed. Some of these methods \cite{pio2021multiverse, zhang2018scalable, liu2017principled} generalize random walks \cite{perozzi2014deepwalk} to the multiplex scenario. They either perform random walks on individual dimensions and then aggregate the results or run random walks that jump from one dimension to another. However, with high-dimensional multiplex graphs, long sequences of nodes are needed to account for all dimensions. Since these methods optimize the embeddings on local patches, it becomes difficult to extract global information. Another line of work uses consensus embeddings to encode information of multiple dimensions \cite{xu2019multi, ni2018co, matsuno2018mell}. A consensus embedding is the representation of a node that includes all dimensions of the graph \cite{xu2019multi}. In particular, these methods leverage regularization to force the embeddings to be similar across dimensions. This strategy is effective in some cases, but in the presence of complementary dimensions, it can hinder the quality of the final representations.

The success of deep neural networks in encoding complex data is effectively aligned with learning appropriate embeddings for multiplex graphs \cite{ma2019multi}. In this regard, Graph Neural Networks (GNNs) \cite{wu2020comprehensive, graphs_deep_learning_survey} extend the deep learning framework to graph-structured data. The most successful attempts to extract latent representations for multiplexes leverage GNNs \cite{ma2019multi, jing2021hdmi}. Most of these methods construct separate embeddings on individual dimensions with GNNs and then aggregate the dimension-specific embeddings into a consensus one using a linear weighted summation. However, these methods consider a single and simple combination. 
A linear aggregation of the dimension-specific embeddings cannot capture complex relations between the graph dimensions. Moreover, it is not clear from previous work how to exploit several aggregations probably because combined representations from different sets of dimensions lead to inconsistent results \cite{boutemine2017mining}. In this context, we argue the existence of relevant hidden dimensions, which can be established hierarchically from the dimension-specific embeddings. Thus, a high-level relation can be seen as a composition of lower-level ones. Unfortunately, existing aggregation strategies cannot handle this compositional aspect. 



\subsection{Motivations}

\begin{figure*}
\centering
\begin{subfigure}[b]{0.40\textwidth}
    \includegraphics[width=\textwidth]{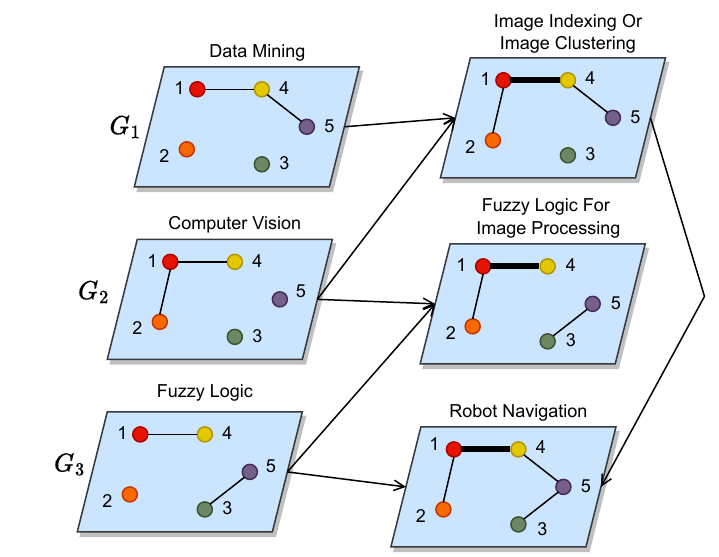}
    \caption{Single vs Multiple Aggregations.}
    \label{fig:single_vs_multiple_example}
\end{subfigure}
\begin{subfigure}[b]{0.49\textwidth}
    \includegraphics[width=\textwidth]{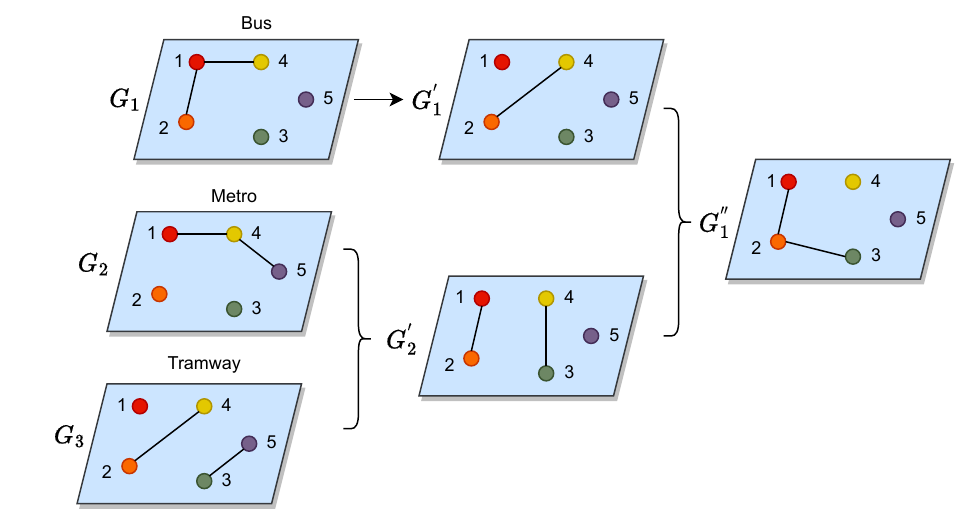}
    \caption{Linear vs Non-Linear Aggregations.}
    \label{fig:latent_structures_example}
\end{subfigure}
\caption{Motivating examples of hierarchical aggregation of multiplex graphs.}
\label{fig:motivations_figure}
\end{figure*}


Multiplex graphs are increasingly characterized by the presence of a high number of dimensions (that is, a large number of links of different types connecting multiple nodes). In this setting, informative and complex latent structures can be hidden across various dimensions. Mining such high-dimensional multiplex graphs continues to pose a challenge to existing methods. In particular, the number of possible combinations grows exponentially with the number of dimensions. Aggregating all dimensions in a \textit{single} \textit{linear} step can cause significant information loss. To explain these problems, we provide two illustrative examples.

\textbf{Single vs Multiple}: We consider the case of a co-authorship multiplex graph, where nodes are the authors of research papers, and edges indicate that two authors have co-written a paper. Edges can be scattered in multiple dimensions such that each dimension represents a publication venue (that is, two authors have co-written a paper in a specific journal or a conference). In this case, different combinations can expose different information about the authors' research domain.

We suppose that two authors have published a paper at a data mining conference and another paper at a computer vision conference. Combining the two dimensions may suggest that the research domain of these authors revolves around ``image indexing or image clustering". If the two authors have also co-written an article in a fuzzy logic journal, then we can infer the research domain ``fuzzy logic for image processing" by combining the data mining and fuzzy logic dimensions. We can see that different combinations can lead to seemingly divergent predictions. Therefore, a single combination at the initial level can cause a significant loss of information by destroying the divergent information. This information can be useful at a higher level of refinement. For example, we can infer from the initial dimension ``fuzzy logic" and the hidden dimension ``image indexing or image clustering" that the authors work on ``robot navigation". We illustrate this example in Figure \ref{fig:single_vs_multiple_example}, where nodes 1 and 4 have a high chance of sharing the aforementioned research domains. From this perspective, we can see that some initial dimensions (e.g., ``fuzzy logic") are more informative when they are used to refine higher-level dimensions (e.g.,``image clustering").

\textbf{Linear vs Non-Linear}: The high-dimensionality in multiplex graphs gives rise to another phenomenon, which is the complex structure associated with information hidden in a large number of dimensions. We consider a multiplex graph that models a transportation network for a given city. Nodes represent locations in the city, and dimensions represent different means of transportation. An edge in a dimension indicates that two locations are directly linked through a transportation mean. We suppose the existence of three dimensions: Bus Lines ($G_1$), Metro ($G_2$), and Tramway ($G_3$). Fig. \ref{fig:latent_structures_example} shows an illustration of some relevant hidden dimensions that can be extracted non-linearly from the initial dimensions.

First, we can extract a relevant latent structure  $G^{'}_1$, which represents the locations that can be reached with two consecutive bus trips. This latent structure can be obtained using a non-linear operation by squaring the adjacency matrix of the Bus Lines dimension $G_1$. Since there is a path $2 \rightarrow 1 \rightarrow 4$ in $G_1$, the latent structure $G^{'}_1$ contains a new link $2 \rightarrow 4$. This link is not present in the initial graph structure $G_1$.
    
Another relevant latent structure $G^{'}_2$ represents the locations that can be accessed by a metro trip followed by a tramway trip. This dimension can be formed by combining the Metro dimension $G_2$ and the Tramway dimension $G_3$ using a non-linear operation. More precisely, the links $1 \rightarrow 4$ in $G_2$ and $4 \rightarrow 2$ in $G_3$ form the link $1 \rightarrow 2$. The link $4 \rightarrow 3$ is formed by $4 \rightarrow 5$ in $G_2$ and $5 \rightarrow 3$ in $G_3$.
    
A third relevant latent structure can be revealed by combining $G^{'}_1$ and $G^{'}_2$. It contains locations that are linked by two bus trips, followed by a metro trip and a tramway trip. This combination forms a new path $1 \rightarrow 2 \rightarrow 3$. This path was not present in any of the original dimensions. Most importantly, the link $2 \rightarrow 3$ cannot be obtained by any linear combination between $G_1$, $G_2$, and $G_3$.

\textbf{Hierarchical Aggregations}: High-dimensional data, such as images or audio, are well-known to be hierarchical in nature \cite{lecun2015deep}. In other words, high-level features are composed of lower-level ones. For example, in image datasets, objects are combinations of motifs, which are themselves combinations of edges. Similarly to high-dimensional data, the complex structure of high-dimensional graphs inherits this compositional aspect. More precisely, there exist high-level hidden dimensions, which can be seen as non-linear compositions of some lower-level dimensions. The two real-world examples presented in Fig. \ref{fig:single_vs_multiple_example} and  \ref{fig:latent_structures_example} show the suitability of hierarchical aggregations to account for this compositional aspect.

\subsection{Contributions}

Due to their hierarchical design, neural networks can construct compositions of low-level features and gradually generate high-level patterns \cite{lecun2015deep, gu2018recent}. Although the state-of-the-art methods harness the deep learning framework to embed the initial dimensions separately, the aggregation step of these methods remains simple and can not discover the compositional structures hidden in the non-linear combinations of dimensions. To capture complex interactions between the initial dimensions, we advocate the adoption of a hierarchical aggregation method. We take inspiration from the deep learning framework to design our aggregation strategy for
high-dimensional multiplex graph embedding, named \methodname (\underline{H}ierarchical \underline{M}ultiplex \underline{G}raph \underline{E}mbedding). 





Our approach hierarchically combines the dimensions of a multiplex graph to construct new dimensions. Each layer of our model computes a more refined multiplex graph with a lower number of dimensions. The last hidden layer produces a one-dimensional graph. The node embeddings are computed by applying a standard GCN on the graph generated by the last hidden layer. Our training process leverages the hidden dimensions computed at each level to gradually extract higher-level ones. This progressive refinement allows to alleviate the information loss caused by the widely used single and linear aggregation step. Since non-linear combinations are computed from previous ones, the proposed approach can uncover complex interactions between the different relations. Finally, to encode globally relevant information present in diverse locations of the graph, we introduce mutual information maximization in the optimization module. The significance of this work can be summarized as follows.

\begin{itemize}
    
    \item \textbf{Methodology:} We propose \methodname, a novel embedding method for high-dimensional multiplex graphs. Our method relies on a hierarchical aggregation strategy to capture complex interactions between the initial dimensions. Progressive refinement of the hidden relations allows to align a large number of divergent and complementary dimensions to a consensus embedding, which in turn alleviates the information loss caused by the widely used single and linear aggregation step.
    

    \item \textbf{Datasets:} To reflect the compositional nature of high-dimensional multiplex graphs in the experiments, we collected from various sources four multiplex graphs with an important number of dimensions. Specifically, we collected two protein-protein interaction graphs, a co-authorship graph, and a movie database graph. We made this data public for future research in the domain.
    
    \item \textbf{Experiments:} We conducted detailed experiments on both synthetic and real high-dimensional multiplex graphs. 
    Specifically, we evaluated \methodname on two downstream tasks: node classification and link prediction. Our results show considerable improvement compared to the state-of-the-art methods for several cases. Furthermore, our ablation study confirms that this improvement is imputed to our hierarchical aggregations.
\end{itemize}



\section{Related work}\label{related_work}

In this section, we review the literature on multiplex graph embedding. Before that, we define two important terms that we use throughout the section. A \textit{dimension-specific node embedding} is the embedding of a node in a given dimension. A \textit{consensus node embedding} is the embedding of a node taking into account all dimensions \cite{xu2019multi}.

\subsection{Random Walk-Based Methods}

Random walk is a popular approach to graph embedding \cite{cui2018survey}, and several methods have been developed to generalize such a method to the multiplex scenario. SMNE \cite{zhang2018scalable} generates random walks in individual dimensions, allowing the method to learn dimension-specific node embeddings. After that, a transformation matrix is applied to obtain consensus node embeddings. MultiVERSE \cite{pio2021multiverse} improves SMNE by performing random walks that can jump from one dimension to another. Additionally, it supports the presence of heterogeneous nodes in the multiplex graph. PMNE \cite{liu2017principled} designs first- and second-order random walks to browse single layers. It uses the Node2Vec \cite{grover2016node2vec} parameters $p$ and $q$ to flexibly traverse the neighbors of the node.

A drawback of random walk-based methods is that they are inherently transductive (that is, they cannot handle the arrival of unobserved nodes). GATNE \cite{cen2019representation} solves this issue by generating training samples using random walks and then computing node embeddings with trainable transformations. In the case of high-dimensional multiplex graphs, random walk-based methods suffer from a major limitation. Indeed, to cover all the graph dimensions, long sequences of random walks must be generated. Since the skip-gram model takes advantage of local patches to optimize representations \cite{mikolov2013distributed}, it can be difficult to capture long-range dependencies in the node sequences. In other words, these methods cannot effectively uncover latent structures spanning multiple dimensions. 
Moreover, random walk-based methods lack the expressive power to model the compositional nature of high-dimensional multiplex graphs.

\subsection{Sharing Information with Regularization}

While learning node embeddings, sharing information between dimensions is essential to capture common and complementary information. 
To this end, a handful of methods rely on regularization. DMNE \cite{ni2018co} uses an auto-encoder to encode each dimension of the multiplex graph. After that, a regularization term is applied to force the dimension-specific embeddings to be similar. This results in a consensus embedding for each node. MELL \cite{matsuno2018mell} follows a similar approach, but for directed multiplex graphs. It uses two embedding vectors for each node: one as a head vector and another as a tail vector. In addition, MELL learns layer-specific representations (i.e., embeddings for entire layers). These representations are used to compute the probability of an edge between two nodes in a specific layer. The optimized loss function is computed from these probabilities, in a similar fashion to the first-order loss of LINE \cite{tang2015line}. MTNE \cite{xu2019multi} also uses a regularization term to encode the different dimensions to the consensus embeddings. In contrast to the other methods, the dimension-specific embeddings are generated from a pretraining stage.


Sharing information with regularization is a simple technique to build consensus embeddings. In fact, most of the regularization-based methods force the dimension-specific embeddings to be similar. However, the used similarity metrics are fixed and simple. Without prior knowledge to systematically identify the best metric, these methods generally underfit the complex interactions between the different dimensions. More precisely, these approaches cannot account for the compositional structures hidden in hierarchical combinations of the initial dimensions. Consequently, the consensus embeddings obtained based on these methods can only contain \textit{low-level} and \textit{common} information. In the presence of complementary dimensions, forcing the embeddings to be similar can cause significant information losses, which in turn hinders the quality of the final representations. Because of these limitations, our approach does not rely on regularization. Instead, it learns to hierarchically combine the multiplex graph dimensions to extract both \textit{common} and \textit{complementary} \textit{high-level} information.



\subsection{Multiplex Graph Neural Networks}

Learning dimension-specific node representations using GNNs and then combining these representations based on simple aggregations (e.g., linear combination, concatenation) is the gold standard for the most recent body of literature. 
For instance, DMGI \cite{park2020dmgi} computes consensus embeddings by assigning attention weights to the node representations of each dimension. 
The training process of DMGI leverages InfoMax (mutual information maximization) \cite{hjelm2018learning} to optimize the latent representations. HDMI \cite{jing2021hdmi} improves on DMGI by defining a higher-order InfoMax mechanism that includes node features. SSDCM \cite{mitra2021semi} applies InfoMax between local node-level representations and global cluster-aware graph summary. 

A limitation of these methods is the loss of information that results from the single and linear aggregation of dimension-specific node embeddings. Different combinations of dimensions can highlight different interactions between dimensions \cite{boutemine2017mining}. This means that some information can only be uncovered by combining specific sets of dimensions. Second, these methods do not account for the compositional nature of high-dimensional multiplex graphs. Consequently, they cannot learn hidden structures that capture complex interactions between different dimensions. On the contrary, our approach hierarchically combines the initial dimensions, making it possible to retrieve latent structures at different levels of the hierarchy. Moreover, our method learns more complex combinations than linear aggregation methods, which we will show in Section \ref{comparison_with_aggregation_methods}.

\section{Proposed Approach}\label{methodology}

Let us first present the notations used throughout the paper and define the problem of multiplex graph embedding.


\textbf{Multiplex Graph}: A multiplex graph is defined as a set of $D$ graphs $G = (G_1, \: G_2, \: \dots, \: G_D)$, where $G_d = (V, \: A_d, \: X)$ is a graph with $N$ nodes from the set $V = \left \{ v_1, v_2, \dots, v_N \right \}$, $A_d \in \{0, 1\}^{N \times N}$ is the adjacency matrix and $X \in \mathbb{R}^{N \times F}$ is the node feature matrix. The nodes and their features are shared across all graphs, while each $G_d$ has its own adjacency matrix. We refer to each graph $G_d$ as a dimension of $G$.

\textbf{Multiplex Graph Embedding}: Given $G$, the goal is to learn an $M$-dimensional vector representation $z_i \in \mathbb{R}^M$ for each node $v_i \in V$, in an unsupervised setting. The vector representations are regrouped in a matrix $Z \in \mathbb{R}^{N \times M}$, which can be used in multiple downstream tasks, such as node classification and link prediction.

\begin{figure*}
\centering
\begin{subfigure}[b]{0.45\textwidth}
    \includegraphics[width=\textwidth]{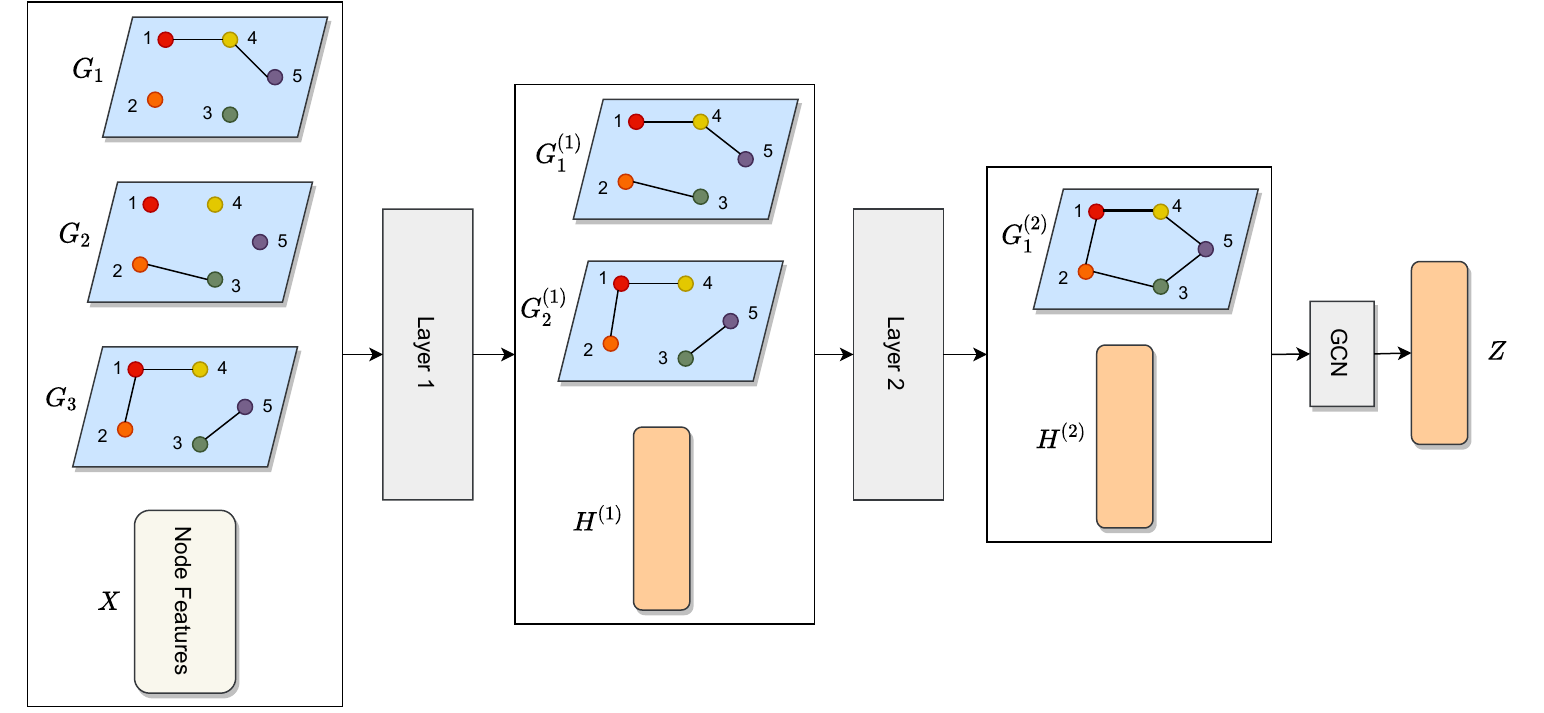}
    \caption{\methodname on a three-dimensional multiplex graph.}
    \label{fig:proposed_approach_figure_a}
\end{subfigure}
\begin{subfigure}[b]{0.54\textwidth}
    \includegraphics[width=\textwidth]{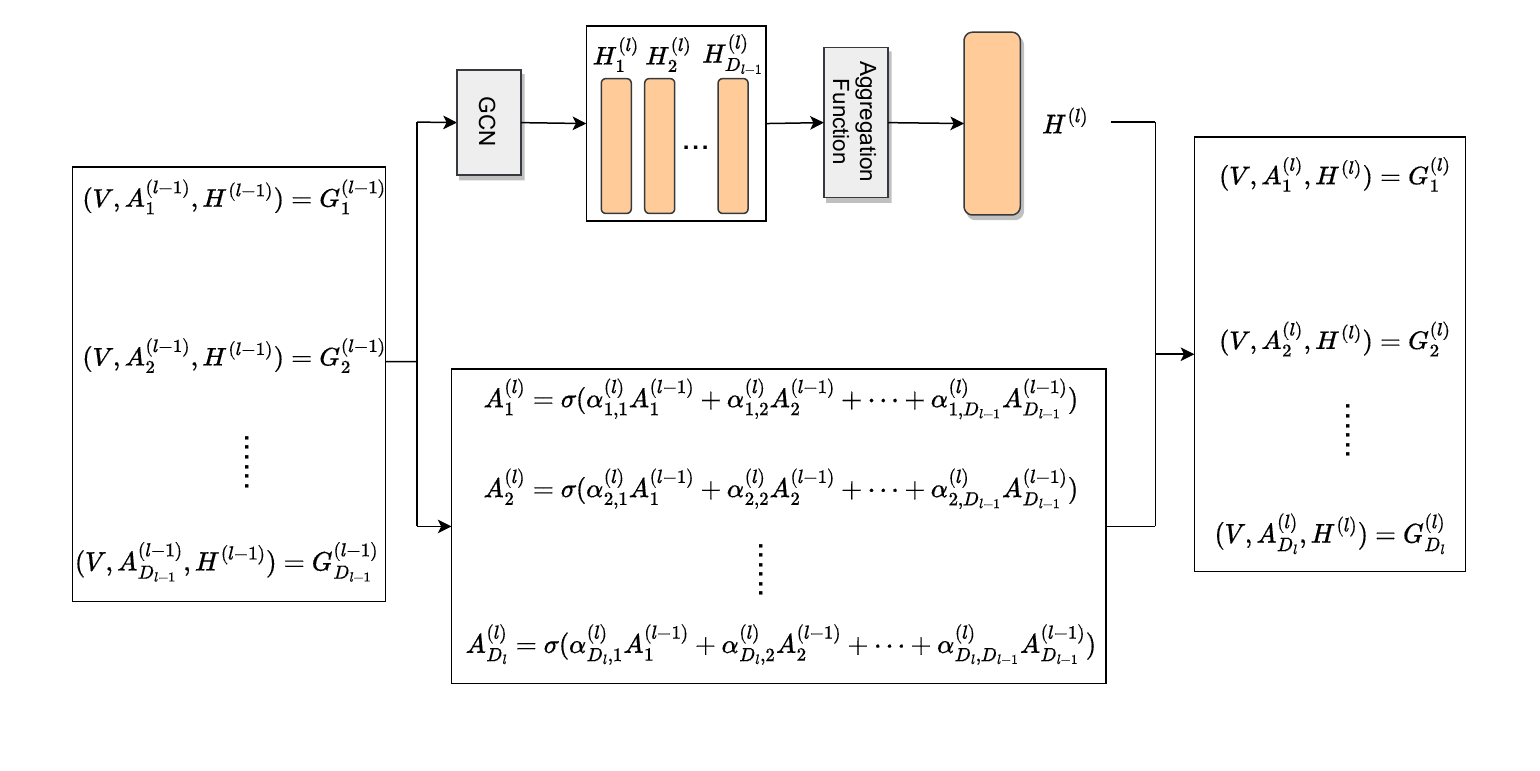}
    \caption{The $l$-th layer of \methodname.}
    \label{fig:proposed_approach_figure_b}
\end{subfigure}
\caption{The architecture of \methodname.}
\label{fig:proposed_approach_figure}
\end{figure*}

\methodname learns node embeddings by computing trainable non-linear combinations of the initial dimensions. More precisely, the graph dimensions are hierarchically combined; that is, each new combination is computed based on combinations from a lower level. Each dimension $G_d$ is represented by its corresponding adjacency matrix $A_d$. New combinations are represented by new adjacency matrices resulting from hierarchical aggregations. 
This process forms a hierarchy of latent dimensions that reflect the compositional aspect of high-dimensional graphs. In fact, hierarchical aggregation aims to encode complex interactions between the initial dimensions in the final embeddings. 
To train the model, we maximize the mutual information between the graph-level representation and the local patches. 

In this section, we give a detailed presentation of \methodname and of the hierarchical aggregation process. After that, we show formally that our hierarchical aggregation strategy can capture more complex patterns than linear aggregation methods such as DMGI \cite{park2020dmgi} and HDMI \cite{jing2021hdmi}. 

\subsection{Overview of \methodname}

Fig. \ref{fig:proposed_approach_figure} depicts the architecture of the proposed approach. \methodname introduces a graph neural network with $L$ layers that can tackle the multiplex case. Unlike previous methods, our approach can learn both latent features and latent graph structures. Specifically, each layer $l$ takes as input a multiplex graph $G^{(l-1)}$, whose attributes are the node embeddings $H^{(l-1)}$ ($G^{(0)} = G$ and $H^{(0)} = X$). Then, this layer outputs a new multiplex graph $G^{(l)}$, whose attributes $H^{(l)}$ and graph structures $A^{(l)}_d$ are computed from $G^{(l-1)}$.

Fig. \ref{fig:proposed_approach_figure_a} illustrates \methodname on a three-dimensional multiplex graph. The first layer computes the node embedding matrix $H^{(1)}$ and transforms the three input dimensions into two dimensions by combining the graph structures of $G_1$ and $G_2$. The second layer computes $H^{(2)}$ by refining $H^{(1)}$ using the newly created dimensions and combines the structures of $G_1^{(1)}$ and $G_2^{(1)}$ to obtain a single graph $G_1^{(2)}$. The final node embeddings $Z$ are computed by running a graph convolutional layer (GCN) on $G_1^{(2)}$.

Multiple transformations are applied inside each layer $l$, as depicted in Fig. \ref{fig:proposed_approach_figure_b}. More precisely, each layer operates in two phases: 1) computing the node embedding matrix $H^{(l)}$; 2) generating hidden dimensions in the form of a multiplex graph $G^{(l)}$. 
In the next subsections, we present in detail how the embeddings are computed and how the hierarchical aggregation strategy refines the initial multiplex graph structures to obtain lower-dimensional multiplex graph structures.

\subsection{Computing Node Embeddings} 


The first phase of the encoding process in each layer $l$ focuses on computing the node embeddings. Given the representation $H^{(l-1)}$ and the graph structures $A^{(l-1)}_d$, a single graph convolutional operation is applied on each dimension $d$ of $G^{(l-1)}$ to obtain new node representations $H^{(l)}_d$:


\begin{equation}
H^{(l)}_d = ReLU(\hat{\Delta}^{-\frac{1}{2}}_d \hat{A}_d^{(l-1)} \hat{\Delta}^{-\frac{1}{2}}_d H^{(l-1)} W^{(l)}_d),
\end{equation}

\noindent where $\hat{A}_d^{(l-1)} = A^{(l-1)}_d + I$, $\hat{\Delta}_d = \text{diag}(\hat{A}_d^{(l-1)} \, \boldsymbol{1}_{N})$, $\boldsymbol{1}_{N} \in \mathbb{R}^{N}$ is a vector of ones, $W^{(l)}_d$ is the weight matrix of the graph convolutional layer, and $ReLU$ is the rectified linear unit activation function. After that, the dimension-specific embeddings $H^{(l)}_d$ are aggregated into a single embedding $H^{(l)}$ as described by the following equations:

\begin{equation}\label{eq:aggregation_formula_1}
    h_n^{(l)} = \sum_{d=1}^{D_{l-1}} \beta^{(l)}_{n, d} \; h^{(l)}_{n, d},
\end{equation}

\begin{equation}\label{eq:aggregation_formula_2}
H^{(l)} = \Big\lVert_{n=1}^N \; h^{(l)}_n,
\end{equation}

\noindent where $h_n^{(l)}$ is the embedding of node $v_n$ generated by the $l^{\text{th}}$ layer, $\beta^{(l)}_{n, d}$ are attention weights \cite{bahdanau2014neural}, $D_{l-1}$ denotes the number of dimensions in $G^{(l-1)}$, and $\big\lVert$ stands for the concatenation operation. 
The matrix $H^{(l)}$ constitutes the node features of the newly generated multiplex graph $G^{(l)}$, which is the input of the next layer of \methodname.

The importance of each dimension in $G^{(l-1)}$ is measured by the attention weights $\beta^{(l)}_{n, d}$, which are computed as follows:

\begin{equation}\label{eq:attention_1}
\hat{\beta}^{(l)}_{n, d} = \text{tanh}(y^{(l)^T}_d \: V^{(l)}_d \: h^{(l)}_{n, d}),
\end{equation}

\begin{equation}\label{eq:attention_2}
\beta^{(l)}_{n, d} = \frac{\hat{\beta}^{(l)}_{n, d}}{\sum_{d^{'} = 1}^{D_{l-1}} \hat{\beta}^{(l)}_{{n, d^{'}}}},
\end{equation}

\noindent where $V^{(l)}_d \in \mathbb{R}^{M \times M}$ and $y^{(l)}_d \in \mathbb{R}^{M}$ are part of the training parameters. The attention weights adjust the contribution of each individual dimension to the representations. In spite of its widespread adoption, 
the attention mechanism is not sufficient to capture complex interactions between the graph dimensions. To address this limitation, the second phase of the encoding process at each layer refines the graph structures.


\subsection{Hierarchical Aggregations} 

The second phase of the encoding process in each layer $l$ focuses on computing the graph structures $A^{(l)}_d$.
As discussed in the motivation subsection, relevant information can be hidden in non-linear compositions of the graph dimensions. To identify this information, the $l^{\text{th}}$ layer computes $D_{l}$ output adjacency matrices from $D_{l-1}$ input adjacency matrices such that $D_{l-1} < D_{l}$.

%

Each output graph structure is computed based on a weighted summation of the input adjacency matrices followed by a non-linear activation function. The weighted summation can uncover hidden paths in the multiplex graph. It also transforms multiple adjacency matrices into a single output matrix. The activation function models non-linear interactions, and stacking multiple \methodname layers can capture more complex interactions. Accordingly, the $j^{\text{th}}$ output adjacency matrix of the $l^{\text{th}}$ layer is computed as follows:

\begin{equation}\label{eq:combination_formula}
    A^{(l)}_{j} = \sigma(\sum_{i=1}^{D_{l-1}} \alpha^{(l)}_{i, j} A^{(l-1)}_{i}),
\end{equation}

\noindent where $\sigma$ is an activation function (we use $ReLU$ in the experiments), and $\alpha^{(l)}_{i, j}$ is the weight associated with the $i^{\text{th}}$ input  and $j^{\text{th}}$ output dimensions. 
Let $\alpha^{(l)} = (\alpha^{(l)}_{i, j}) \in \mathbb{R}^{D_{l-1} \times D_{l}}$ be the matrix that contains the combination weights. Since $\alpha^{(l)}$ is a trainable matrix, \methodname learns combinations of adjacency matrices that maximize the mutual information objective function. Note that before computing Eq. \eqref{eq:combination_formula}, the weights $\alpha^{(l)}_{i, j}$ are normalized with a softmax function:

\begin{equation}
    \alpha^{(l)}_{i, j} = \frac{\exp{(\alpha^{(l)}_{i, j})}}{\sum_{k = 1}^{D_{l-1}} \exp{(\alpha^{(l)}_{k, j})}}.
\end{equation}

The $l^{\text{th}}$ layer learns and computes multiple combinations, which are then used by the layer $l+1$ to improve the node embeddings $H^{(l)}$ and compute other combinations. This process forms a hierarchy of adjacency matrices as illustrated in Fig. \ref{fig:proposed_approach_figure_a}. 
In addition, since each layer increments on the embeddings of the previous layer, the final representation $Z$ contains information about both the input graph $G$ and the intermediate graphs $G^{(l)}$. Moreover, by applying an activation function on the output dimensions and stacking multiple layers, \methodname can uncover non-linear latent structures hidden in the input multiplex graph dimensions. 


To tackle the high-dimensionality of real-world multiplex graphs, the proposed solution gradually reduces the number of dimensions of the input at each layer. When the number of dimensions is high, aggregating all embeddings in a single step is ineffective, as it can cause a significant loss of information. \methodname addresses this issue by aggregating the embeddings multiple times on several intermediate multiplex graphs. We show in the next subsection that this hierarchical aggregation process can model more complex patterns than linear aggregation.

\subsection{\methodname vs Linear Aggregation Methods}\label{comparison_with_aggregation_methods}

Most current approaches to multiplex graph embedding rely on linear aggregation to extract node representations \cite{park2020dmgi}, \cite{jing2021hdmi}. More specifically, they compute node embeddings on individual dimensions and then aggregate them using a weighted summation. Attention weights are used to account for the relevance of each dimension. In this section, we show that hierarchical aggregations can model more complex patterns than linear aggregation approaches.

For readability, we ignore the attention weights in the following equations and suppose that all graph convolutional operations have the same weight matrix $W$. Moreover, we ignore the activation functions to clearly exhibit the combinations that are formed by the different methods. For illustration purposes, consider a two-dimensional multiplex graph in the input. The following equation summarizes the embedding module of a linear aggregation method with two layers:
\begin{equation}\label{eq:aggregration_based_formula}
\begin{split}
    Z & = A_1 \, (A_1 X W) \, W + A_2 \, (A_2 X W) \, W, \\
      & = (A_1^2 + A_2^2) \, X \, W^2.
\end{split}
\end{equation}

We can see that $A_1^2$ and $A_2^2$ are summed to compute the node embedding matrix $Z$. On the other side, a two-layer \methodname embedding module on the same input graph can be summarized as follows:

\begin{equation}\label{eq:hdme_formula}
\begin{split}
    Z & = (\alpha_1 A_1 + \alpha_2 A_2) \, H^{(1)} \, W, \\
      & = (\alpha_1 A_1 + \alpha_2 A_2) \, (A_1 X W + A_2 X W) \, W, \\
      & = (\alpha_1 A_1 + \alpha_2 A_2) \, (A_1 + A_2) \, X \, W^2, \\
      & = (\alpha_1 A_1^2 + \alpha_1 A_1 A_2 + \alpha_2 A_2 A_1 + \alpha_2 A_2^2) \, X \, W^2, \\
      & = (\alpha_1 A_1^2 + (\alpha_1 + \alpha_2) A_1 A_2 + \alpha_2 A_2^2) \, X \, W^2.
\end{split}
\end{equation}

\methodname can learn more complex patterns than linear aggregation methods. Using hierarchical aggregations, both $A_1^2$ and $A_2^2$ are leveraged, while also introducing $A_1 A_2$. Moreover, the weights $\alpha_1$ and $\alpha_2$ make it possible to consider the importance of each term. The term $A_1 A_2$ may reveal edges that cannot be retrieved by considering $A_1$ and $A_2$ individually, or by combining them linearly. For example, as illustrated in Fig. \ref{fig:latent_structures_example}, the combination of two adjacency matrices (specifically, in this figure, the product of the matrices) forms relevant paths between nodes that were not previously connected. 

In practice, given a high-dimensional multiplex graph, several \methodname layers can be stacked to construct even more complex combinations. 
In the experiments section, we show that the additional expressive power of \methodname improves the performance on downstream tasks compared to linear aggregation methods on both synthetic and real-world data.

\subsection{Training Algorithm}\label{loss_function_section}

\methodname is trained based on a self-supervised strategy \cite{graph_selfsupervised_survey}, which involves the learning of general-purpose node embeddings. Therefore, the trained model can be used to perform multiple downstream tasks. 
Specifically, inspired by Deep Graph Infomax (DGI) \cite{velickovic2019deep}, we employ a loss based on mutual information maximization \cite{hjelm2018learning}. The training procedure maximizes the mutual information between the graph-level representation and the local patches from the set $\{z_1, z_2, \dots, z_N\} $. The graph-level representation $s$ is a summary of the entire graph $G$ and is obtained by aggregating all the node embeddings $z_i$:

\begin{equation}
    s = \frac{1}{N} \sum_{i = 1}^{N} z_i.
\end{equation}

A discriminator $\mathscr{D}$ is used as a proxy to maximize the intractable mutual information function. The discriminator takes as input a patch-summary pair and is trained to compute the probability score assigned to the pair. To this end, we select positive pairs, which consist of the graph summary $s$ and samples from $Z = \textit{HDME}(G)$. 
On the other hand, negative pairs are formed from $s$ and samples from $\hat{Z} = \textit{HDME} (\hat{G})$. The graph $\hat{G}$ is a corrupted version of the input graph that can be generated in multiple ways, such as swapping nodes and removing or adding links. In this work, we randomly shuffle the node features to obtain a corrupted version $\hat{X}$ and use this version as the feature matrix of $\hat{G}$. The discriminator function $\mathcal{D}$ is implemented based on bilinear scoring:


\begin{equation}
    \mathscr{D}(h_i, s) = \text{Sigmoid}(h_i \, Q \, s),
\end{equation}

\noindent where $Q \in \mathbb{R}^{M \times M}$ is a parameter matrix, and $M$ is the size of node embeddings. We minimize the binary cross-entropy loss $\mathcal{L}$ to train the discriminator and generate the node embeddings as described by:

\begin{equation}\label{eq:loss_function}
    \mathcal{L} = \sum_{i=1}^{N} \log \mathscr{D}(z_i, s) + \sum_{j=1}^{N} \log ( 1 - \mathscr{D}(\hat{z}_j, s)).
\end{equation}

\begin{algorithm}[t]
\caption{\textit{\methodname}}
\label{algo:method_algo_1}
\begin{algorithmic}[1]

\REQUIRE multiplex graph $G$, 
embedding dimension $M$, number of layers $L$, number of iterations $T$.
\ENSURE  Node embeddings $Z$.

\FOR{$epoch \gets 1$ to $T$}
    \STATE $\hat{X} \gets \text{Shuffle}(X)$
    \STATE $\hat{G} \gets G$, but replace $X$ with $\hat{X}$.
    \STATE $Z \gets \text{Encode-Multiplex-Graph}(G)$
    \STATE $\hat{Z} \gets \text{Encode-Multiplex-Graph}(\hat{G})$
    \STATE $s \gets \frac{1}{N} \sum_{i = 1}^{N} z_i$
    \STATE $\mathcal{L} \gets \sum_{z_i \in Z} \log \mathscr{D}(z_i, s) + \sum_{\hat{z}_j \in \hat{Z}} \log ( 1 - \mathscr{D}(\hat{z}_j, s))$
    \STATE Update $\{\alpha^{(l)}, W_d^{(l)}, V_d^{(l)}, y_d^{(l)}\}$ to minimize $\mathcal{L}$ with Adam.
\ENDFOR

\STATE \textbf{return} Z.
\end{algorithmic}
\end{algorithm}

\begin{algorithm}[t]
\caption{\textit{Encode-Multiplex-Graph}}
\label{algo:method_algo_2}
\begin{algorithmic}[1]

\REQUIRE multiplex graph $G$, number of layers $L$.
\ENSURE node embeddings $H^{(L)}$.

\STATE $G^{(0)} \gets G$
\STATE $H^{(0)} \gets X$
\FOR{$l \gets 1$ to $L$}
    \STATE $G^{(l)}, H^{(l)} \gets \text{Hierarchical-Aggregation}(G^{(l-1)}, H^{(l-1)})$
\ENDFOR
\STATE \textbf{return} $H^{(L)}$.
    
\end{algorithmic}
\end{algorithm}

\begin{algorithm}[t]
\caption{\textit{Hierarchical-Aggregation}}
\label{algo:method_algo_3}
\begin{algorithmic}[1]

\REQUIRE multiplex graph $G^{(l-1)}$, node embeddings  $H^{(l-1)}$. 
\ENSURE  multiplex graph $G^{(l)}$, node embeddings $H^{(l)}$. 

\STATE $H^{(l)} \gets \text{Phase-One}(G^{(l-1)}, \, H^{(l-1)})$
\STATE $G^{(l)} \gets \text{Phase-Two}(G^{(l-1)})$
\STATE \textbf{return} $G^{(l)}, \, H^{(l)}$. \newline
\STATE \textbf{Function} Phase-One$(G^{(l-1)}, \, H^{(l-1)})$
\begin{ALC@g}
    \FOR{$d \gets 1$ to $D_{l-1}$}
        \STATE $H^{(l)}_d \gets \text{GCN}(H^{(l-1)}, \, A^{(l-1)}_d)$
    \ENDFOR
    \STATE Compute $\beta^{(l)}$ according to Eq. \eqref{eq:attention_1} and Eq. \eqref{eq:attention_2}.
    \STATE $h_n^{(l)} = \sum_{d=1}^{D_{l-1}} \beta^{(l)}_{n, d} \; h^{(l)}_{n, d}$
    \STATE $H^{(l)} = \Big\lVert_{n=1}^N \; h^{(l)}_n$
    \STATE \textbf{return} $H^{(l)}.$ \newline
\end{ALC@g}

\STATE \textbf{Function} Phase-Two$(G^{(l-1)})$
\begin{ALC@g}
    \FOR{$j \gets 1$ to $D_{l}$}
        \STATE $A^{(l)}_{j} \gets \sigma(\sum_{i}^{D_{l-1}} \alpha^{(l)}_{i, j} A^{(l-1)}_{i})$
        \STATE $G^{(l)}_j \gets (V, \, A^{(l)}_{j}, \, H^{(l)})$
    \ENDFOR
    \STATE $G^{(l)} \gets (G_1^{(l)}, \, G_2^{(l)}, \, \dots, \, G_{D_l}^{(l)})$
    \STATE \textbf{return} $G^{(l)}$.
\end{ALC@g}
\end{algorithmic}
\end{algorithm}

Algorithms \ref{algo:method_algo_1}, \ref{algo:method_algo_2} and \ref{algo:method_algo_3} summarize the proposed approach. We train the model for $T$ iterations and update the parameters $\{\alpha^{(l)}, W_d^{(l)}, V_d^{(l)}, y_d^{(l)}\}$ by minimizing the loss $\mathcal{L}$ using the Adam optimizer. The time complexity of \methodname is $\mathcal{O}(T L D (\mathcal{E} M + N M^2 + \mathcal{E} D))$, where $N$ is the number of nodes, $D$ the number of input dimensions, $L$ the number of embedding layers, $M$ the size of node embeddings, and $\mathcal{E}$ is the maximum number of edges in the graph dimensions.  The memory complexity of our approach is $\mathcal{O}(L D (\mathcal{E} + N M + M^{2}+ 2 D^{2}))$.


\section{Experiments}\label{experiments}


In this section, we evaluate the suitability of \methodname for high-dimensional multiplex graph embedding. First, we perform experiments on synthetically generated data. After that, we compare \methodname with various multiplex graph embedding methods on real-world data for two downstream tasks: link prediction and node classification. Finally, we present an ablation study followed by visualizations of the embeddings and combination weights $\alpha^{(l)}$.


\begin{table*}
\centering
\begin{tabular}{ c|c|c|c|c|c } 
\hline
Dataset & \# Dimensions & \# Nodes & \# Edges & \# Training data & \# Classes \\ 
\hline
\hline
BIOGRID & 15 & 4,211 & 280 979 & 252 & 4 \\
\hline
DBLP-Authors & 10 & 5,124 & 33 250 & 307 & 4 \\
\hline
IMDB & 8 & 3000 & 224 984 & 173 & 3 \\
\hline
STRING-DB & 7 & 4,083 & 4 923 554 & 244 & 3 \\ 
\hline
\end{tabular}
\caption{Real-world data statistics.}
\label{table:data_statistics}
\end{table*}

\subsection{Experimental Setup}

\subsubsection{Datasets}
Due to the scarcity of real-world labeled high-dimensional multiplex graphs, we collected from various sources four datasets to assess the suitability of \methodname. Table \ref{table:data_statistics} summarizes the statistics of these multiplex graphs. The datasets and the implementation of \methodname are 
on Github \footnote[2]{\url{https://github.com/abdouskamel/HMGE}}. 

\textbf{BIOGRID:} We collected a multiplex graph from the BIOGRID database of proteins \cite{oughtred2021biogrid}. Nodes represent proteins, and edges represent protein-protein interactions. Protein-protein interactions can be inferred by various protocols, for example, by using the biochemical effect of one protein on another or from X-ray Crystallography at the atomic level \cite{oughtred2021biogrid}. The dimensions of the multiplex constitute different inference protocols. Node classification labels represent the species from which the proteins have been extracted.
    
\textbf{DBLP-Authors:} This dataset represents a coauthorship graph. We used AMiner \cite{tang2016aminer} to collect it. The nodes are the authors of research papers, and an edge between two authors indicates that they have co-written a paper. Dimensions represent various conferences and journals. For example, if two authors are connected in dimension $d_1$, they have co-written a paper in the conference or journal $d_1$. We use the authors' research areas as classification labels. Note that we end up with a multilabel classification problem (i.e., a node can belong to more than one class).
    
\textbf{IMDB:} It is a multiplex graph extracted from IMDB \footnote{\url{https://www.imdb.com/interfaces/}}. Nodes are movies, and an edge between two movies indicates that a person has participated in both movies. Different dimensions represent different roles: actors, directors, producers, etc. We use the movie genre as classification labels.
    
\textbf{STRING-DB:} This is another protein-protein interaction graph collected from STRING-DB \cite{szklarczyk2019string}. It is similar to BIOGRID in terms of node, edge, and dimension semantics. The difference with BIOGRID is in the classification labels. In this dataset, the labels represent protein families instead of animal species.


\subsubsection{Compared Algorithms}
Our comparison focuses on methods designed for multiplex graph embedding. 

\textbf{CrossMNA \cite{chu2019cross}:} This method defines node embeddings as a combination of intra-vectors (i.e., dimension-specific embeddings) and inter-vectors (i.e., consensus embeddings). It then optimizes a loss based on edge reconstruction.

\textbf{MultiVERSE \cite{pio2021multiverse}:} It is a random walk-based method. It learns embedding by performing random walks on the supra-adjacency matrix of the multiplex graph (i.e., walks can jump from one dimension to another). After that, the random walks are turned into a similarity matrix from which node embeddings are then generated. 

\textbf{GATNE \cite{cen2019representation}:} This method runs random walks on each dimension of the graph and then generates new training samples based on nodes' proximity. It then computes embeddings with a trainable linear combination mixed with attention. Finally, it optimizes a loss function based on graph reconstruction.
    
\textbf{DMGI \cite{park2020dmgi}:} It is a linear aggregation method that learns individual node embeddings in each dimension by maximizing mutual information. After that, embeddings are aggregated using attention scores. A regularization term is employed to fine-tune the representations.
    
\textbf{HDMI \cite{jing2021hdmi}:} It is an extension of DMGI that defines a high-order mutual information loss. On each dimension, it maximizes the mutual information between a graph summary, local patches, and node features.

\textbf{\textbf{SSDCM \cite{mitra2021semi}}:} Similar to DMGI and HDMI, this method is based on mutual information maximization. However, structure-aware global graph representations are leveraged to optimize the InfoMax loss. This is achieved by a cluster-aware strategy for graph summary generation.


\subsubsection{Evaluation Protocol}
To evaluate \methodname, we perform both link prediction and node classification. We use the unsupervised loss in Eq. \eqref{eq:loss_function} to learn the node embeddings for both tasks. Then, for node classification, we run logistic regression on the embeddings and evaluate the results with F1-Macro and F1-Micro. For link prediction, we use the inner product activated with a Sigmoid function to calculate the missing links scores: $\text{Sigmoid}(ZZ^T)$. The results are then evaluated with the area under the ROC curve (AUC-ROC) and average precision (AP). Note that for link prediction, we randomly remove $10\%$ of the links for the sake of evaluation. 


\subsubsection{Parameter Settings} For our method, we set the size of node embeddings to $64$ (except on BIOGRID, where we set it to $32$, as it leads to better results) and the number of layers to $2$. We use Adam as an optimizer with a learning rate of $0.001$ and a weight decay of $10^{-5}$. We train for $2,000$ epochs and adopt early stopping with a patience of $100$. 

\subsection{Experiments on Synthetic Data}

We first compare \methodname with DMGI and HDMI on generated data. Through this evaluation, we show that when the graph dimensions increase, the classification accuracy of the competitors rapidly decreases. In contrast, the accuracy of \methodname is significantly less affected by the increase in the number of dimensions.

\subsubsection{Data Generation}

To synthesize a multiplex graph with $N$ nodes, $D$ dimensions and $K$ classes, we run a stochastic block model (SBM) \cite{holland1983stochastic} $D$ times. Each run generates a dimension of the graph along with node labels specific to this dimension. The SBM proceeds as follows: (1) Each node has a probability $p_i$ to belong to class $c_i$. We set $K = 2$ (binary classification) with $p_1 = p_2$; (2) For every pair of nodes $v_i$ and $v_j$, such that $v_i$ is in class $c_i$ and $v_j$ is in class $c_j$, the edge $(v_i, v_j)$ has a probability $p_{i, j}$. We use two probabilities $p_{in}$ and $p_{out}$: the probabilities of an edge between two nodes of the same class and of different classes, respectively.

After running SBM $D$ times, we obtain $D$ adjacency matrices each with its own classification labels. We use a voting mechanism across the dimensions to get a single global label for each node. For example, if $v$ is in class $c_1$ in dimension $G_1$, in class $c_2$ in dimension $G_2$, and in class $c_1$ in dimension $G_3$, we assign $v$ to class $c_1$. This process creates a multiplex graph whose dimensions contribute to the node labels. In this context, an effective embedding method should capture as much information as possible from every dimension.


In total, we generate $9$ synthetic multiplex graphs with 3,000 nodes and with the following number of dimensions: $\{3, 7, 11, 15, 21, 25, 31, 35, 41\}$. We set $p_1 = p_2 = 0.5$, $p_{in} = 0.05$, and $p_{out} = 0.01$. We generate binary classes and use the classification accuracy for evaluation.

\subsubsection{Evaluation Results}

\begin{figure}
    \centering
    \includegraphics[width=0.5\textwidth]{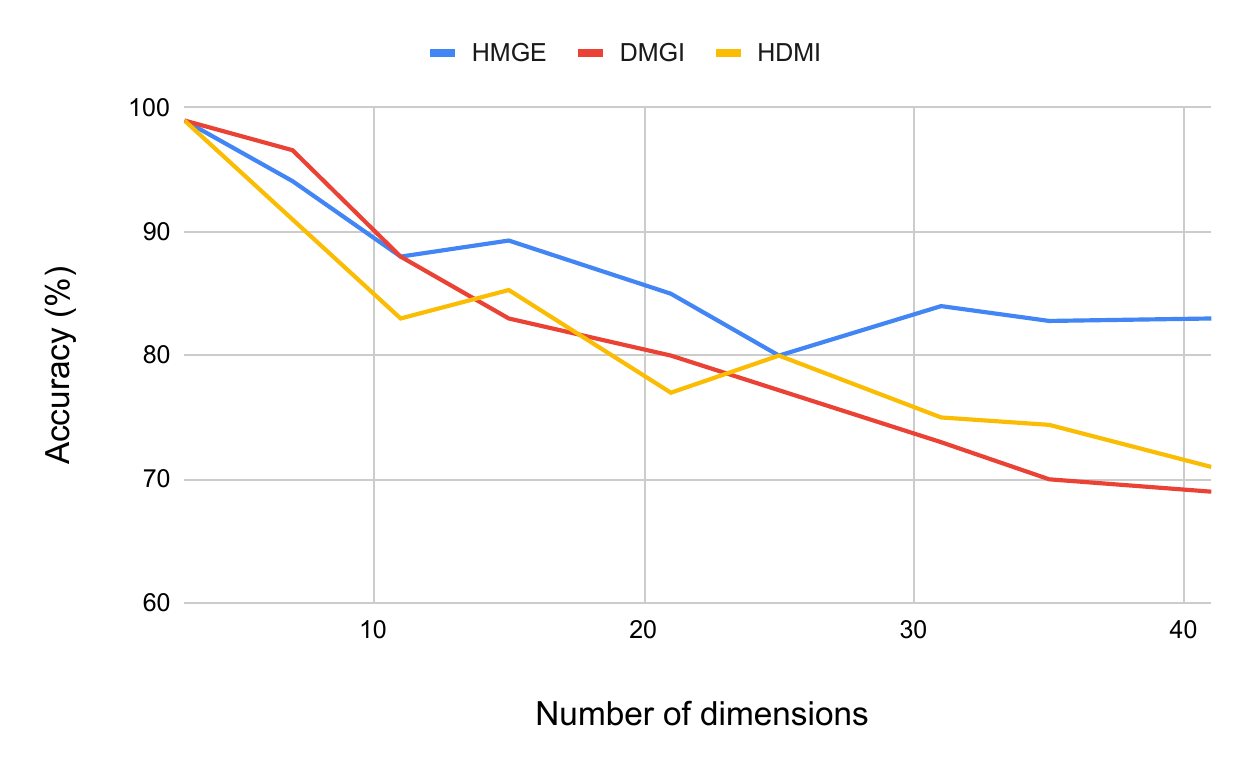}
    \caption{Node classification results on synthetic data.}
    \label{fig:synthetic_data_chart}
\end{figure}

Fig. \ref{fig:synthetic_data_chart} shows the node classification results on the synthetic datasets. We can see that the accuracy of DMGI and HDMI significantly decreases as the number of dimensions increases, going from a near-perfect classification when $D = 3$ to below $70\%$ when $D = 41$. 
As the number of dimensions increases, the global node labels become more and more different from the dimension-specific labels, making predicting the correct global node labels more difficult. On the other hand, the accuracy of \methodname drops at a considerably slower rate, reaching $83\%$ when $D = 41$. 

Note that the competing algorithms have near-perfect accuracy when $D = 3$. This is because each dimension has a simple structure consisting of two highly dense regions connected by a sparse region. When $D$ is small, predicting the node labels from this simple structure is straightforward, but the prediction becomes more difficult as $D$ increases. These results illustrate the difficulty of effectively encoding high-dimensional multiplex graphs.

Furthermore, the results suggest that competitors considered in the comparison do not fully exploit the information provided by each dimension. 
This aspect exhibits a serious limitation of state-of-the-art methods, particularly common embedding techniques that use linear aggregation. 
\methodname alleviates this issue by leveraging hierarchical aggregations. 

\subsection{Experiments on Real-World Data}

\begin{table*}
\centering
\begin{tabular}{ c|c|c|c|c|c|c|c|c  } 
\hline
Dataset & \multicolumn{2}{|c|}{BIOGRID} & \multicolumn{2}{|c|}{DBLP-Authors} & \multicolumn{2}{|c|}{IMDB} & \multicolumn{2}{|c}{STRING-DB} \\

\hline 
Metrics & AUC & AP & AUC & AP & AUC & AP & AUC & AP \\
\hline
CrossMNA & \underline{50.95} & \underline{50.48} & 53.17 & 51.63 & 50.9 & 50.45 & 50.19 & 50.09\\
MultiVERSE & 50.3 & 50.15 & 54.54 & 52.38 & 48.34 & 49.15 & 50 & 50 \\
GATNE & 39.82 & 42.53 & 43.57 & 44.34 & 37.54 & 41.05 & 48.15 & 47.89 \\
SSDCM & 32.14 & 40.52 & \underline{62.74} & 58.54 & \underline{53.84} & \underline{54.48} & 50.4 & 50.2 \\
DMGI & 42.33 & 44.16 & 50 & 50 & 52.84 & 52.66 & \underline{54.62} & 52.69 \\
HDMI & 43.91 & 46.01 & 60.64 & \underline{59.36} & 50.95 & 50.76 & 54.46 & \underline{53.96} \\
\hline
\methodname & \textbf{71.76} & \textbf{70.17} & \textbf{67.31} & \textbf{67.21} & \textbf{57.42} & \textbf{55.77} & \textbf{65.46} & \textbf{62.84} \\
\hline
\end{tabular}
\caption{Results on link prediction. Best in bold and second best underlined.}
\label{table:link_prediction}
\end{table*}

\begin{table*}
\centering
\begin{tabular}{ c|c|c|c|c|c|c|c|c  } 
\hline
Dataset & \multicolumn{2}{|c|}{BIOGRID} & \multicolumn{2}{|c|}{DBLP-Authors} & \multicolumn{2}{|c|}{IMDB} & \multicolumn{2}{|c}{STRING-DB} \\

\hline
Metrics & F1-Macro & F1-Micro & F1-Macro & F1-Micro & F1-Macro & F1-Micro & F1-Macro & F1-Micro \\
\hline
CrossMNA & 97.04 & 96.98 & \textbf{63.52} & 70.08 & 34.42 & 34.33 & 33.26 & 72.44 \\
MultiVERSE & \underline{98.53} & \underline{98.53} & 57.9 & 60.72 & 41.3 & 41.3 & 46.68 & 47.75 \\
GATNE & 98.43 & 98.49 & \underline{58.12} & 71.34 & \underline{42.52} & \underline{42.31} & 70.1 & 72.11 \\
SSDCM & 11.22 & 28.42 & 57.67 & \underline{71.70} & 24.78 & 33.59 & 61.13 & 65.84 \\
DMGI & 50.98 & 51.96 & 54.49 & 62.36 & 38.2 & 38.2 & 65.61 & 67.62  \\
HDMI & 62.9 & 64.46 & 57.1 & 70.8 & 38.9 & 39.5 & \underline{72.03} & \underline{73.94} \\
\hline
\methodname & \textbf{98.75} & \textbf{98.77} & 57.52 & \textbf{71.76} & \textbf{43.02} & \textbf{43.16} & \textbf{80.33} & \textbf{82.08} \\
\hline
\end{tabular}
\caption{Results on node classification. Best in bold and second best underlined.}
\label{table:node_classification}
\end{table*}

%

In this section, we evaluate \methodname on real-world high-dimensional multiplex graphs collected from various sources (Table \ref{table:data_statistics}). We perform two downstream tasks: link prediction and node classification.

\subsubsection{Link Prediction}

Table \ref{table:link_prediction} shows a comparison between the proposed approach and state-of-the-art methods on the link prediction task. We can see that most compared algorithms have AUC and AP scores below or near 50\%. 
These methods are not well-suited for link prediction on high-dimensional multiplex graphs for several reasons. On the one hand, MultiVERSE and GATNE require long sequences of random walks to consider all dimensions. Moreover, these methods can not capture long-range dependencies in the node sequences. On the other hand, SSDCM, DMGI, and HDMI are limited by the single and linear aggregation of dimension-specific node embeddings. Therefore, they fail to capture the compositional relations between the initial dimensions.

We can also see from Table \ref{table:link_prediction} that \methodname yields better results than the compared algorithms. In particular, our method outperforms the state-of-the-art models by a significant margin in several cases. 
For instance, the difference in the link prediction task between \methodname and the most competitive method on BIOGRID (i.e., CrossMNA) is higher than 20\% in terms of AUC. Furthermore, \methodname achieves AUC and AP scores over 70\% on BIOGRID, and close to 70\% on DBLP-Authors. These two datasets have the highest number of dimensions among the other ones. Our results substantiate the effectiveness of \methodname in encoding high-dimensional multiplex graphs. Unlike previous methods, our approach stands out by its ability to perform hierarchical aggregations, which are introduced to capture compositional interactions between the initial dimensions. 




\subsubsection{Node Classification}

Table \ref{table:node_classification} shows the evaluation results on node classification. \methodname outperforms HDMI, DMGI, and SSDCM, which are based on mutual information maximization and linear aggregation. Particularly, these methods are less competitive on BIOGRID, which is the dataset with the highest number of dimensions. For instance, the difference in node classification performance between \methodname and DMGI on BIOGRID is higher than 40\% in terms of F1-Macro and F1-Micro. As the number of dimensions increases, the linear aggregation strategy becomes less effective. We reported similar results on synthetic data. In the high-dimensional setting, informative and complex latent structures can be hidden across various dimensions. These latent structures can be established hierarchically from the dimension-specific embeddings.

We can also see from Table \ref{table:node_classification} that our model yields better results than random walk-based methods (GATNE and MultiVERSE). These methods use the skip-gram model, which has limited capacity to capture long-range dependencies in the node sequences. Unlike these methods, \methodname leverages hierarchical aggregations to capture the compositional interactions between the initial dimensions in the final node embeddings.

\begin{table*}
\centering
\begin{tabular}{ c|c|c|c|c|c|c|c|c } 
\hline
Dataset & \multicolumn{2}{|c|}{BIOGRID} & \multicolumn{2}{|c|}{DBLP-Authors} & \multicolumn{2}{|c|}{IMDB} & \multicolumn{2}{|c}{STRING-DB} \\ 
\hline
Metrics & AUC & AP & AUC & AP & AUC & AP & AUC & AP \\
\hline
\methodname (First Ablation) & 47.16 & 47.17 & \underline{56.02} & \underline{53.76} & 41.73 & 43.70 & 59.75 & 58.23 \\
\methodname (Second Ablation) & \underline{66.48} & \underline{63.38} & 47.91 & 49.05 & \underline{52.45} & \underline{51.28} & \underline{62.64} & \underline{60.02} \\
\hline
\methodname & \textbf{71.76} & \textbf{70.17} & \textbf{67.31} & \textbf{67.21} & \textbf{57.42} & \textbf{55.77} & \textbf{65.46} & \textbf{62.84} \\
\hline
\end{tabular}
\caption{Ablation study on the task of link prediction. Best in bold and second best underlined.}
\label{table:ablation_link_prediction}
\end{table*}

\begin{table*}
\centering
\begin{tabular}{ c|c|c|c|c|c|c|c|c } 
\hline
Dataset & \multicolumn{2}{|c|}{BIOGRID} & \multicolumn{2}{|c|}{DBLP-Authors} & \multicolumn{2}{|c|}{IMDB} & \multicolumn{2}{|c}{STRING-DB} \\ 
\hline
Metrics & F1-Macro & F1-Micro & F1-Macro & F1-Micro & F1-Macro & F1-Micro & F1-Macro & F1-Micro \\
\hline
\methodname (First Ablation) & 91.07 & 91.26 & 56.74 & 70.07 & \underline{39.18} & \underline{40.97} & 61.95 & 66.69 \\
\methodname (Second Ablation) & \underline{95.87} & \underline{96.03} & \underline{57.35} & \underline{70.35} & 21.29 & 34.75 & \underline{79.61} & \underline{81.45} \\
\hline
\methodname & \textbf{98.75} & \textbf{98.77} & \textbf{57.52} & \textbf{71.76} & \textbf{43.02} & \textbf{43.16} & \textbf{80.33} & \textbf{82.08} \\
\hline
\end{tabular}
\caption{Ablation study on the task of node classification. Best in bold and second best underlined.}
\label{table:ablation_node_classification}
\end{table*}

\subsection{Ablation Study}

In this section, we perform two ablation experiments to show the significance of our contributions: ablation of the whole hierarchical aggregation mechanism, and ablation of the combination weights.

\subsubsection{First Ablation (Hierarchical Aggregation Mechanism)}

We evaluate the performance of \methodname without using the hidden layers that generate the latent multiplex graphs. When the number of hidden layers is zero, our training process amounts to learning the embeddings on each dimension with GCNs and then aggregating them linearly to a single representation. The goal is to illustrate the benefits of hierarchical aggregations to  high-dimensional multiplex graph embedding.

The first rows of Tables \ref{table:ablation_link_prediction} and \ref{table:ablation_node_classification} show the results obtained after performing the first ablation, respectively on the tasks of link prediction and node classification. We can see that the hierarchical aggregation mechanism significantly improves the link prediction and node classification results. The progressive refinement of the hidden relations allows to align a large number of divergent and complementary dimensions to a consensus embedding, which in turn alleviates the information loss caused by the widely used single and linear aggregation step. Note that the optimal number of layers depends on the dataset and should be selected through empirical tests.

 
\subsubsection{Second Ablation ($\alpha$ Weights)}

We evaluate the performance of \methodname without using the $\alpha$ weights. More precisely, we compare between using Eq. \eqref{eq:combination_formula} to compute the non-linear combinations, and using the same formula without weights:
\begin{equation}
    A^{(l)}_{j} = \sigma(\sum_{i=1}^{D_{l-1}} A^{(l-1)}_{i}).
\end{equation}

The second rows of Tables \ref{table:ablation_link_prediction} and \ref{table:ablation_node_classification} show the results obtained after performing the second ablation, respectively on the tasks of link prediction and node classification. We can see that the $\alpha$ weights grant the trained model more flexibility to improve the link prediction and node classification results. These weights are necessary to adjust the importance of each latent dimension. Otherwise, all the dimensions would contribute equally to the generated latent structures. To identify relevant non-linear combinations, some dimensions must be attenuated or even discarded using the trainable $\alpha$ weights. 

\subsection{Hyper-parameters Sensitivity}

\begin{figure}
    \centering
    \includegraphics[width=0.49\textwidth]{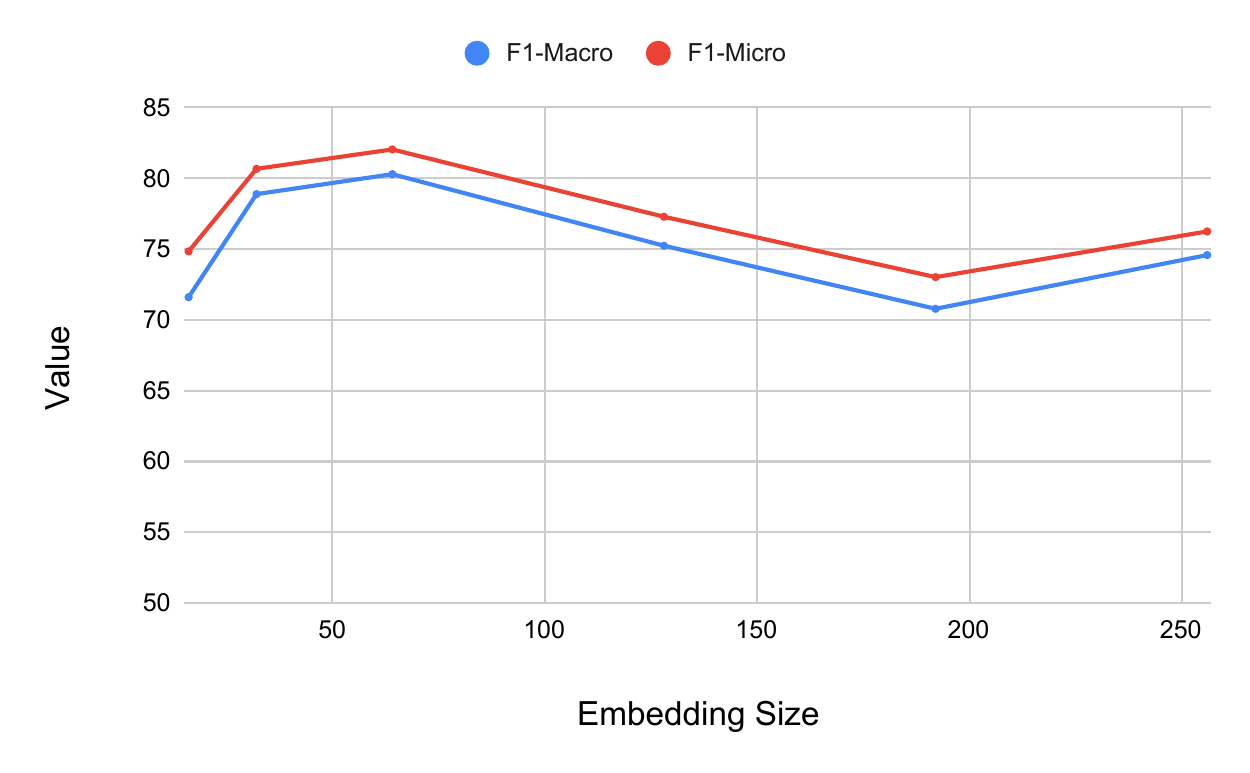}
    \caption{Sensitivity to the embeddings size on STRING-DB}
    \label{fig:embedding_size_sensitivity}
\end{figure}

Fig. \ref{fig:embedding_size_sensitivity} shows the variations of F1-macro and F1-micro scores with respect to the node embedding size $M$. 
We can see that the variations remain small as the scores fluctuate within the range of 71\% and 82\%. Since \methodname yields consistent results for a wide range of values, we can conclude that our approach is robust with respect to the node embedding size.

\begin{figure*}
    \centering
    \begin{subfigure}[b]{0.32\textwidth}
        \includegraphics[width=\textwidth]{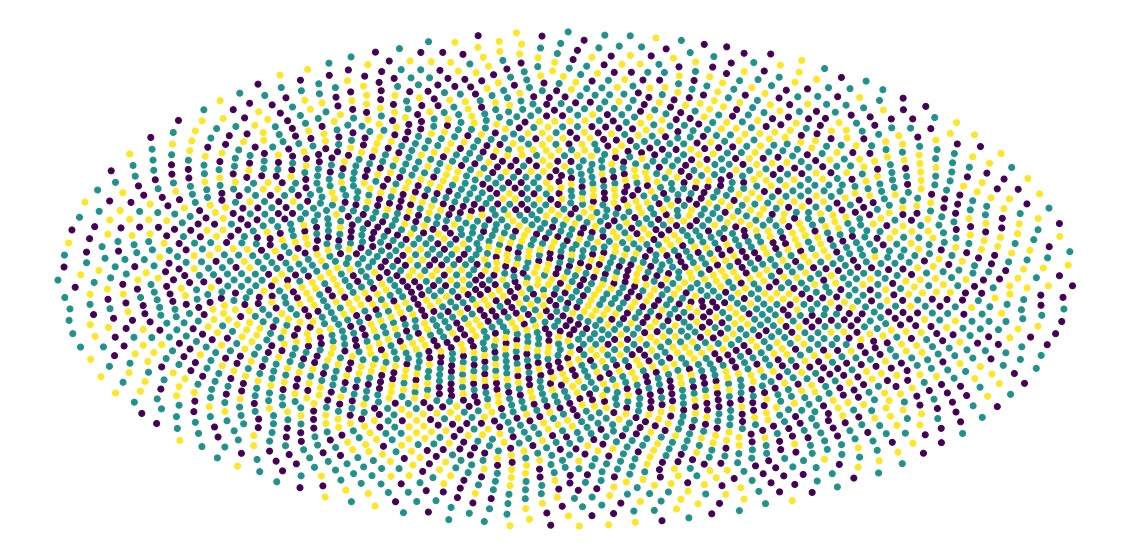}
        \caption{Layer 0.}
        \label{fig:viz_embs_a_stringdb}
    \end{subfigure}
    \begin{subfigure}[b]{0.32\textwidth}
        \includegraphics[width=\textwidth]{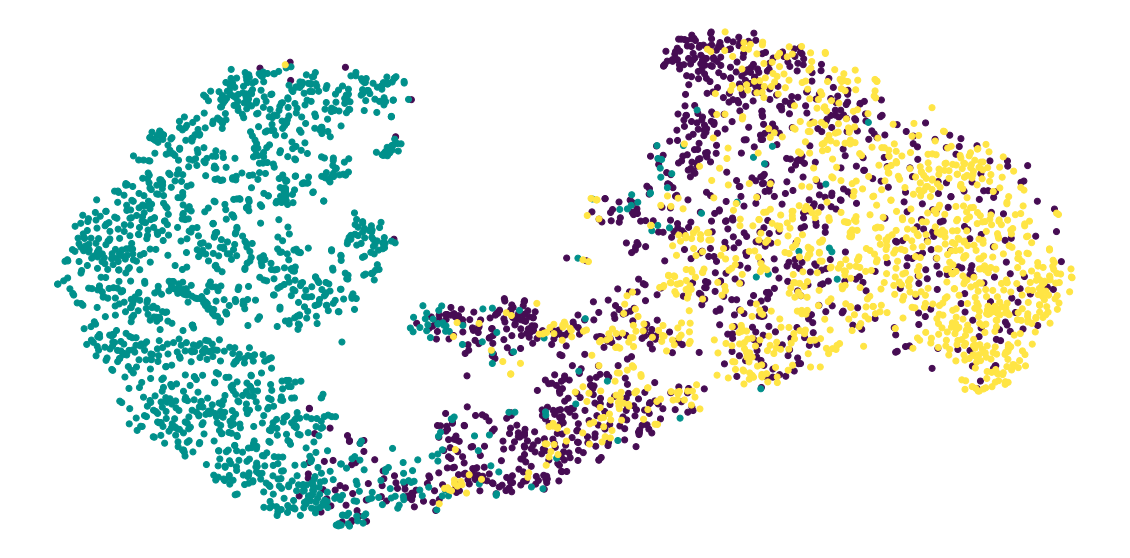}
        \caption{Layer 1.}
        \label{fig:viz_embs_b_stringdb}
    \end{subfigure}
    \begin{subfigure}[b]{0.32\textwidth}
        \includegraphics[width=\textwidth]{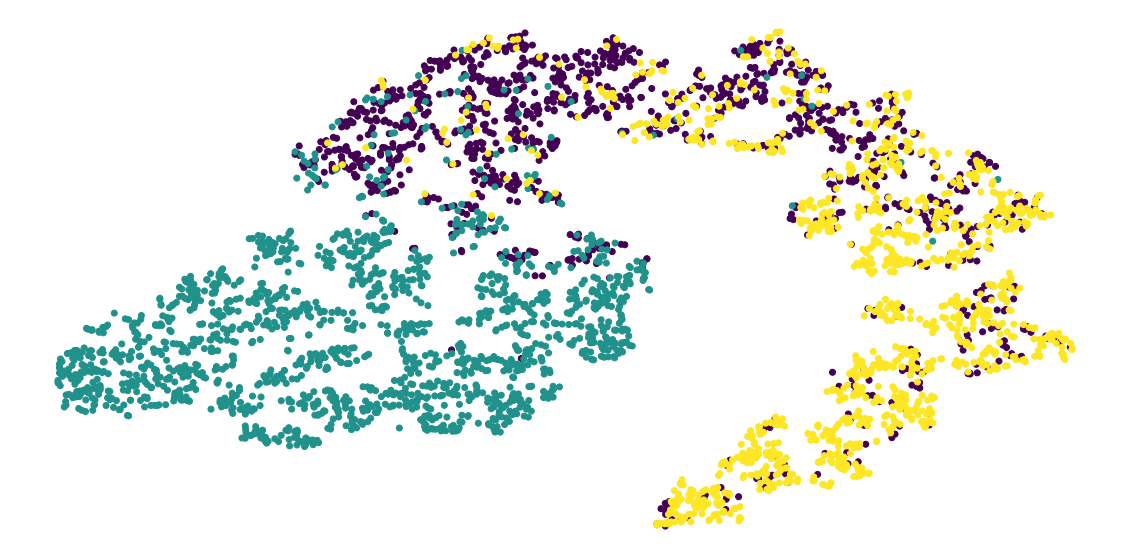}
        \caption{Layer 2.}
        \label{fig:viz_embs_c_stringdb}
    \end{subfigure}
    \caption{Visualization of the embeddings learned at different layers of \methodname on STRING-DB.}
    \label{fig:visualize_embeddings_stringdb}
\end{figure*}

\begin{figure*}
    \centering
    \begin{subfigure}[b]{0.32\textwidth}
        \includegraphics[width=\textwidth]{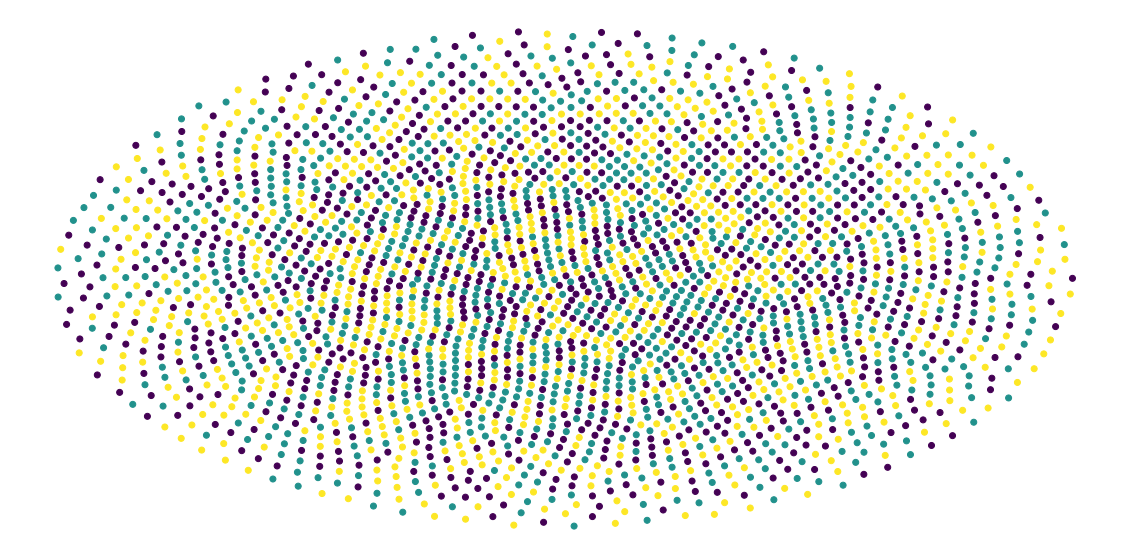}
        \caption{Layer 0.}
        \label{fig:viz_embs_a_imdb}
    \end{subfigure}
    \begin{subfigure}[b]{0.32\textwidth}
        \includegraphics[width=\textwidth]{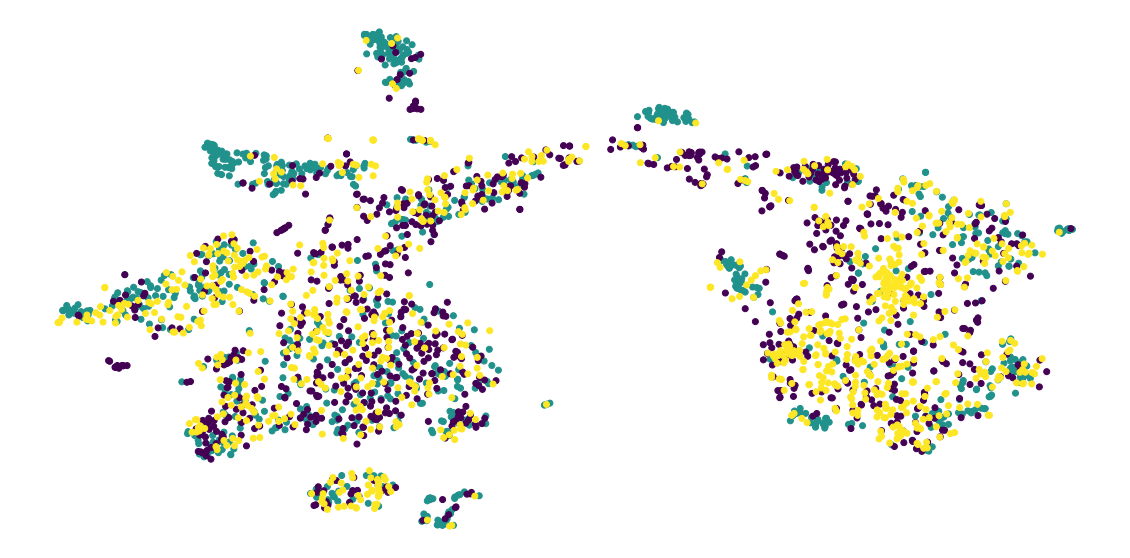}
        \caption{Layer 1.}
        \label{fig:viz_embs_b_imdb}
    \end{subfigure}
    \begin{subfigure}[b]{0.32\textwidth}
        \includegraphics[width=\textwidth]{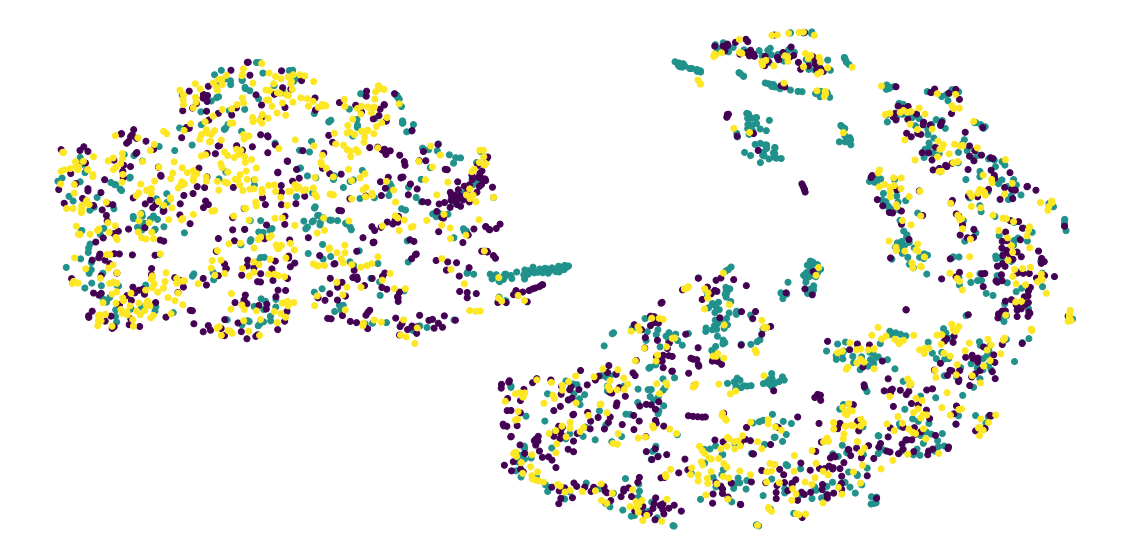}
        \caption{Layer 2.}
        \label{fig:viz_embs_c_imdb}
    \end{subfigure}
    \caption{Visualization of the embeddings learned at different layers of \methodname on IMDB.}
    \label{fig:visualize_embeddings_imdb}
\end{figure*}

\subsection{Visualizations}

In this subsection, we show qualitative results to give additional insights into the inner working of \methodname. 

\subsubsection{Node Embedding}

Fig. \ref{fig:visualize_embeddings_stringdb} and \ref{fig:visualize_embeddings_imdb} show 2D T-SNE visualizations of the embedding matrix $H^{(l)}$ at multiple layers of \methodname, respectively on STRING-DB and IMDB. Different colors represent different classes. Generally, we can see that each layer improves upon the separation between the different classes. Note that layer 0 represents the input layer.

On STRING-DB, the classes cannot be linearly separated at layer 0. At layer 1, the blue class is well separated from the rest of the classes, but the purple and yellow classes are still tangled. Layer 2 improves the geometric configuration by pushing away samples from the yellow class from samples from the purple one. We conclude that, for STRING-DB, one layer is sufficient to identify the blue class. A second layer is necessary to better separate the purple and yellow classes. This illustrates the benefits of a hierarchical aggregation approach to multiplex graph embedding. On IMDB, layer 1 forms the foundation for partitioning the nodes into 2 communities. Layer 2 improves on this separation. 
However, the classes are still not linearly separable. 
Since the model is trained without ground-truth labels, the representations may not align with all downstream tasks.

\subsubsection{Combination Weights}

\begin{figure}
    \centering
    \includegraphics[width=0.45\textwidth]{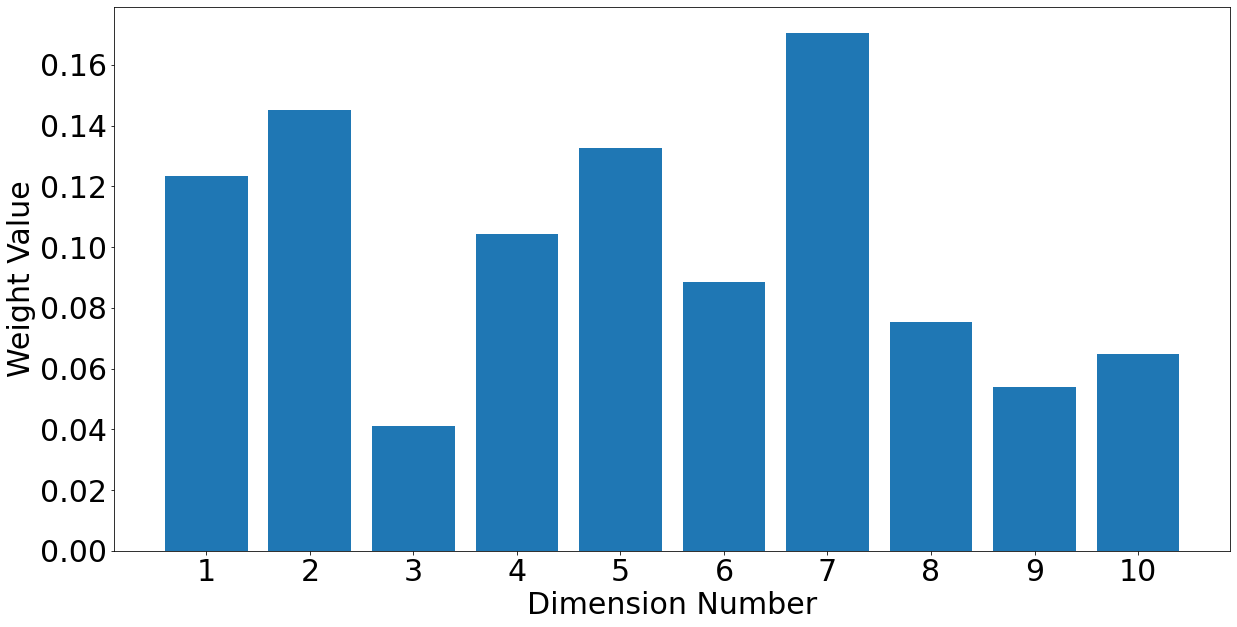}
    \caption{Visualization of the combination weights $\alpha^{(1)}$ on DBLP-Authors.}
    \label{fig:visualize_weights}
\end{figure}

Fig. \ref{fig:visualize_weights} shows a visualization of the combination weights $\alpha^{(1)}$ of the first layer on DBLP-Authors. The horizontal axis shows the different dimensions of the first layer. The vertical axis shows the weight of each dimension. 

We can see that dimensions 3, 8, 9, and 10 have been assigned small weights compared to the other dimensions. In the ablation study, we showed that the combination weights have an important impact on the quality of the final embeddings. In particular, the results of link prediction on DBLP-Authors significantly drop when not using combination weights. The visual results confirm that some dimensions are less relevant than the others for link prediction.

\section{Conclusion}\label{conclusion}

In this paper, we propose \methodname, a novel approach for embedding high-dimensional multiplex graphs. 
In the high-dimensional setting, informative latent structures are hidden in non-linear combinations of the initial dimensions. In this context, traditional methods based on linear aggregation struggle to achieve suitable consensus embeddings. To tackle these issues, our approach hierarchically encodes the input multiplex graph to low-dimensional node representations using hierarchical aggregations. At each level of the hierarchy, hidden dimensions are formed by computing non-linear combinations of the input dimensions. This process gradually generates lower-dimensional multiplex graphs and identifies relevant latent structures hidden in the initial graph. 
Experiments on synthetic and real-world datasets show that our approach outperforms the state-of-the-art methods in multiplex graph embedding. The ablation study confirms the relevance and effectiveness of hierarchical aggregations for high-dimensional multiplex graph embedding.

\ifCLASSOPTIONcompsoc
  \section*{Acknowledgments}
\else
  \section*{Acknowledgment}
\fi

This work has been supported by Research Grants from the Natural Sciences and Engineering Research Council of Canada (NSERC). 

\ifCLASSOPTIONcaptionsoff
  \newpage
\fi

\balance
\bibliographystyle{IEEEtran}
\bibliography{HMGE.bbl}

\end{document}